# Space Efficiency of Propositional Knowledge Representation Formalisms

**Marco Cadoli**                                                        CADOLI@DIS.UNIROMA1.IT
*Dipartimento di Informatica e Sistemistica*
*Università di Roma "La Sapienza"*
*Via Salaria 113, I-00198, Roma, Italy*

**Francesco M. Donini**                                                DONINI@DIS.UNIROMA1.IT
*Politecnico di Bari*
*Dipartimento di di Elettrotecnica ed Elettronica*
*Via Orabona 4, I-70125, Bari, Italy*

**Paolo Liberatore**                                                   LIBERATO@DIS.UNIROMA1.IT
**Marco Schaerf**                                                      SCHAERF@DIS.UNIROMA1.IT
*Dipartimento di Informatica e Sistemistica*
*Università di Roma "La Sapienza"*
*Via Salaria 113, I-00198, Roma, Italy*

## Abstract

We investigate the *space efficiency* of a Propositional Knowledge Representation (PKR) formalism. Intuitively, the space efficiency of a formalism $F$ in representing a certain piece of knowledge $\alpha$, is the size of the shortest formula of $F$ that represents $\alpha$. In this paper we assume that knowledge is either a set of propositional interpretations (models) or a set of propositional formulae (theorems). We provide a formal way of talking about the relative ability of PKR formalisms to compactly represent a set of models or a set of theorems. We introduce two new compactness measures, the corresponding classes, and show that the relative space efficiency of a PKR formalism in representing models/theorems is directly related to such classes. In particular, we consider formalisms for nonmonotonic reasoning, such as circumscription and default logic, as well as belief revision operators and the stable model semantics for logic programs with negation. One interesting result is that formalisms with the same time complexity do not necessarily belong to the same space efficiency class.

## 1. Introduction

During the last years a large number of formalisms for knowledge representation (KR) have been proposed in the literature. Such formalisms have been studied from several perspectives, including semantic properties, and computational complexity. Here we investigate *space efficiency*, a property that has to do with the minimal size needed to represent a certain piece of knowledge in a given formalism. This study is motivated by the fact that the same piece of knowledge can be represented by two formalisms using a different amount of space. Therefore, all else remaining the same, a formalism could be preferred over another one because it needs less space to store information.

The definition of space efficiency, however, is not simple. Indeed, a formalism may allow several different ways to represent the same piece of knowledge. For example, let us assume that we want to represent the piece of knowledge "today is Monday". In Propositional

Logic we may decide to use a single propositional variable *monday*. The fact that today is Monday can be represented by the formula *monday*, but also by the formula $\neg\neg monday$, as well as *monday* $\wedge$ (*rain* $\vee$ $\neg rain$), because all formulae of the Propositional Logic that are logically equivalent to *monday* represent exactly the same information.

In Propositional Logic, we should consider the shortest of the equivalent formulae used to represent the information we have. The same principle can be applied to a generic formalism: if it allows several formulae to represent the same information, then we only take into account the shortest one. Therefore, we say that the space efficiency of a formalism $F$ in representing a certain piece of knowledge $\alpha$ is the size of the shortest formula of $F$ that represents $\alpha$. Space efficiency —also called *succinctness* or *compactness*— of a formalism is a measure of its ability in representing knowledge in a small amount of space.

In this paper we focus on propositional KR (PKR) formalisms. We do not give a formal definition of which formalisms are propositional and which one are not: intuitively, in a propositional formalism, quantifications are not allowed, and thus the formulae are syntactically bounded to be formed only using propositional connectives, plus some other kind of nonclassical connectives (for instance, negation in logic programs, etc.).

So far, we have not discussed what knowledge represents. A possible way to think of a piece of knowledge is that it represents all facts that can be inferred from it. In other words, knowing something is the same as knowing everything that can be logically implied. The second way — which is in some cases more natural — is to think of a piece of knowledge as the set of states of the world that we consider possible.

In a more formal way, we say that knowledge is represented either by a set of propositional interpretations (those describing states of the world we consider plausible) or a set of formulae (those implied from what we know). Consequently, we focus on both reasoning problems of model checking and theorem proving. The following example shows that we can really think of knowledge in both ways.

**Example 1** *We want to eat in a fast food, and want to have either a sandwich or a salad (but not both), and either water or coke (but not both).*

*In Propositional Logic, each choice can be represented as a model, and the following models represent all possible choices (models are represented by writing down only the letters mapped to true).*

$$A = \{\{sandwich, water\}, \{sandwich, coke\}, \{salad, water\}, \{salad, coke\}\}$$

*For representing the set of choices we can use formulae instead of models. In this case, we write down a set of formulae whose models represent exactly the allowed choices, as follows.*

$$\begin{aligned} C \;\; = \;\; & (sandwich \vee salad) \wedge (\neg sandwich \vee \neg salad) \wedge (sandwich \rightarrow \neg salad) \;\wedge \\ & (water \vee coke) \wedge (\neg water \vee \neg coke) \wedge (\neg coke \rightarrow water) \end{aligned}$$

*Actually, we can get rid of redundancies, and end up with the following formula.*

$$F = (sandwich \vee salad) \wedge (\neg sandwich \vee \neg salad) \wedge (water \vee coke) \wedge (\neg water \vee \neg coke)$$

*More formally, F represents the set of models A, because for each interpretation I, I ∈ A holds if and only if I ⊨ F. The formula F also represents the set of formulae C, because Cn(F) = Cn(C), where Cn(.) is the function that gives the set of all conclusions that can be drawn from a propositional formula.*

## 1.1 State of the Art

A question that has been deeply investigated, and is related to space efficiency, is the possibility of translating a formula expressed in one formalism into a formula expressed in another formalism (under the assumption, of course, that these formulae represent the same knowledge).

In most cases, the analysis is about the possibility of translating formulae from different formalisms to Propositional Logic (PL). For example, Ben-Eliyahu and Dechter (1991, 1994) proposed a translation from default logic to PL, and a translation from disjunctive logic programs to PL, while Winslett (1989) introduced a translation from revised knowledge bases to PL, and Gelfond, Przymusinska, and Przymusinskyi (1989) defined a translation from circumscription to PL.

All the above translations, as well as many other ones in the literature, lead to an exponential increase of the size of the formula, in the worst case. When the best known translation yields a formula in the target formalism which has exponential size w.r.t. the formula in the source formalism, a natural question arising is whether such exponential blow up is due to the specific translation, or is intrinsic of the problem. For example, although all proposed translations from default logic to PL lead to the exponential blow up, we cannot conclude that all possible translations suffer from this problem: it could be that a polynomial translation exists, but it has not discovered so far.

Some works have focussed on the question of whether this kind of exponential increase in the size is intrinsic or not. Cadoli, Donini, and Schaerf (1996) have shown that many interesting fragments of default logic and circumscription cannot be expressed by polynomial-time fragments of PL without super-polynomially increasing the size of formulae. It has been proved that such a super-polynomial increase of size is necessary when translating unrestricted propositional circumscription (Cadoli, Donini, Schaerf, & Silvestri, 1997) and most operators for belief revision into PL (Cadoli, Donini, Liberatore, & Schaerf, 1999; Liberatore, 1995).

Gogic and collegues (1995) analyzed the relative succinctness of several PKR formalisms in representing sets of models. Among other results, they showed that skeptical default logic can represent sets of models more succinctly than circumscription.

Kautz, Kearns, and Selman (1995) and Khardon and Roth (1996, 1997) considered representations of knowledge bases based on the notion of *characteristic model*, comparing them to other representations, e.g., based on clauses. They showed that the representation of knowledge bases with their characteristic models is sometimes exponentially more compact than other ones, and that the converse is true in other cases.

However, all the above results are based on specific proofs, tailored to a specific reduction, and do not help us to define equivalence classes for the space efficiency of KR formalisms. In a recent paper (Cadoli, Donini, Liberatore, & Schaerf, 1996b), a new complexity measure for decision problems, called *compilability*, has been introduced. In the present paper we show how this new measure can be directly used to characterize the space efficiency of PKR formalisms. We emphasize methodological aspects, expressing in a more general context many of the results presented before.

## 1.2 Goal

The notion of polynomial time complexity has a great importance in KR (as well as many other fields of computer science), as problems that can be solved in polynomial time are to be considered easy, from a computational point of view.

The notion of *polynomial many-one reducibility* also has a very intuitive meaning when applied to KR: if there exists a polynomial many-one reduction from one formalism to another one, then the time complexity of reasoning in the two formalisms is comparable. This allows to say, e.g., that inference in PL is coNP-complete, i.e. it is one of the hardest problems among those in the complexity class coNP.

As a result, we have a formal tool for comparing the difficulty of reasoning in two formalisms. What is missing is a way for saying that one formalism is able to represent the same information in less space.

**Example 2** *We consider again the lunch scenario of the previous example. We show that we can reduce the size of the representation using circumscription instead of Propositional Logic. In PL, the knowledge of the previous example was represented by the formula F:*

$$F = (sandwich \lor salad) \land (\neg sandwich \lor \neg salad) \land (water \lor coke) \land (\neg water \lor \neg coke)$$

*The set of models of this formula is A, and the models of A are exactly the minimal models of the formula $F_c$ defined as follows.*

$$F_c = (sandwich \lor salad) \land (water \lor coke)$$

*By the definition of circumscription (McCarthy, 1980) it holds that F is equivalent to $CIRC(F_c; \{sandwich, salad, water, coke\}, \emptyset, \emptyset)$. Note that $F_c$ is shorter than F. If this result can be proved to hold for arbitrary sets of models, we may conclude that circumscription is more space efficient than Propositional Logic in representing knowledge expressed as sets of models.*

Our goal is to provide a formal way of talking about the relative ability of PKR formalisms to compactly represent information, where the information is either a set of models or a set of theorems. In particular, we would like to be able to say that a specific PKR formalism provides "one of the most compact ways to represent models/theorems" among the PKR formalisms of a specific class.

## 1.3 Results

We introduce two new compactness measures (*model* and *theorem compactness*) and the corresponding classes (model-C and thm-C, where C is a complexity class like P, NP, coNP, etc.). Such classes form two hierarchies that are isomorphic to the polynomial-time hierarchy (Stockmeyer, 1976). We show that the relative space efficiency of a PKR formalism is directly related to such classes. In particular, the ability of a PKR formalism to compactly represent sets of models/theorems is directly related to the class of the model/theorem hierarchy it belongs to. Problems higher up in the model/theorem hierarchy can represent sets of models/theorems more compactly than formalisms that are in lower classes.

This classification is obtained through a general framework and not by making direct comparisons and specific translations between the various PKR formalisms. Furthermore, our approach also allows for a simple and intuitive notion of completeness for both model and theorem hierarchies. This notion precisely characterizes both the relation between formalisms at different levels, and the relations between formalisms at the same level. An interesting result is that two PKR formalisms in which model checking or inference belong to the same time complexity class may belong to different compactness classes. This may suggest a criterion for choosing between two PKR formalisms in which reasoning has the same time complexity—namely, choose the more compact one. Also, two PKR formalisms may belong to the same theorem compactness class, yet to different model compactness classes. This stresses the importance of clarifying whether one wants to represent models or theorems when choosing a PKR formalism.

### 1.4 Outline

In the next section we introduce the notation and the assumptions that we adopt in this work. In Section 3 (Compilability) we briefly recall some notions on non-uniform computation that are important for what follows and we recall the basic definitions of compilability classes (Cadoli et al., 1996b). In Section 4 (Reductions) we describe the constraints we impose on reductions, while in Section 5 (Space Efficiency) we introduce our compactness classes. In Section 6 (Applications) we actually compare many known PKR formalisms using our framework. Finally, in Section 7 (Related Work and Conclusions) we compare our work with other proposals presented in the literature and draw some conclusions.

## 2. Notations and Assumptions

In this section we define what knowledge bases and formalisms are. Since we want to consider formalisms that are very different both in syntax and in semantics, we need very general definitions. Let us consider, as a base case, the formalism of propositional calculus. Formally, we can assume that it is composed of three parts:

1. a *syntax*, which is used to define the well-formed formulae;

2. a *proof theory*, which allows for saying when a formula follows from another one; and

3. a *model-theoretic semantics*, which establishes when a model satisfies a formula.

The syntax is defined from a finite alphabet of propositional symbols $L = \{a, b, c, \ldots\}$, possibly with subscripts, and the usual set of propositional connectives $\wedge$, $\vee$, $\neg$.

In terms of knowledge representation, the proof theory can be seen as a way for extracting knowledge from a knowledge base. For example, if our knowledge base is $a \wedge c$, then the fact $a \vee b$ holds. We can thus say that the formula $a \vee b$ is part of the knowledge represented by $a \wedge c$.

In some cases, we want knowledge bases to represent models rather than sets of formulas. An *interpretation* for an alphabet of propositional variables $L$ is a mapping from $L$ in {true, false}. The *model-theoretic semantics* of the propositional calculus is the usual way of extending an interpretation for $L$ to well-formed formulas.

Let us now extend such definition to generic formalisms: a formalism is composed of a syntax, a proof theory, and a model-theoretic semantics.

We remark that each formalism has its own syntax: for instance, default logic includes a ternary connective $\dot{-}$ for denoting default rules, while logic programming has a special unary connective $not()$, and so on. A knowledge base of a formalism $F$ is simply a well-formed formula, according to the syntax of the formalism.

Each formalism has its own proof theory as well. The proof theory of a formalism $F$ is a binary relation $\vdash_F$ on the set of knowledge bases and formulae. Intuitively, $FB \vdash_F \phi$ means that $\phi$ is a consequence of the knowledge base $KB$, according to the rules of the formalism $F$. As a result, the set of formulae $\phi$ that are implied by a knowledge base $KB$ is exactly the knowledge represented by $KB$.

The base of a comparison between two different formalisms is a concept of equivalence, allowing for saying that two knowledge bases (of two different formalisms) represent the same piece of knowledge. Since the knowledge represented by a knowledge base is the set of formulas it implies, we have to assume that the syntax of these formulae is the same for all formalisms. Namely, we always assume that the formulae implied by a knowledge base are well-formed formulae of the propositional calculus. In other words, each formalism has a syntax for the knowledge bases: however, we assume that the proof theory relates knowledge bases (formulae in the syntax of the formalism) with propositional formulae. So, while writing $KB \vdash_F \phi$, we assume that $KB$ is a knowledge base in the syntax of $F$, while $\phi$ is a propositional formula.

This allows for saying that two knowledge bases $KB_1$ and $KB_2$, expressed in two different formalisms $F_1$ and $F_2$, represent the same piece of knowledge: this is true when, for any propositional formula $\phi$ it holds $KB_1 \vdash_{F_1} \phi$ if and only if $KB_2 \vdash_{F_2} \phi$.

The *model-theoreric semantics of a formalism* is a relation $\models_F$ between propositional models and knowledge bases. In this case, we assume a fixed alphabet $L$, thus the set of all interpretations is common to all formalisms. When a model $M$ and a knowledge base $KB$ are in the relation, we write $M \models_F KB$. Intuitively, this means that the model $M$ supports the piece of knowledge represented by $KB$.

We remark that some formalisms, e.g. credolous default logic (Reiter, 1980), have a proof theory, but do not have a model-theoretic semantics. It is also possible to conceive formalisms with a model-theoretic semantics but no proof theory. When both of them are defined, we assume that they are related by the following formula:

$$KB \vdash_F \phi \quad \text{iff} \quad \forall I \; . \; I \models KB \text{ implies } I \models \phi$$

Regarding the proof theory of formalisms, we only consider formulae that are shorter than the knowledge base, that is, we assume that the knowledge represented by a knowlegde base $KB$ is the set of formulae $\phi$ such that $KB \vdash_F \phi$, and the size of $\phi$ is at most the size of $KB$. This is done for two reasons: first, formulas that are larger than $KB$ are likely to contain large parts that are actually independent from $KB$; second, we can give technicals result in a very simple way by using the compilability classes introduced in the next section.

**Assumption 1** *We consider only formulae whose size is less than or equal to that of the knowledge base.*

All formalisms we consider satisfy the right-hand side distruibutivity of conjunction, that is, $KB \vdash_F \phi \wedge \mu$ if and only if $KB \vdash_F \phi$ and $KB \vdash_F \mu$. The assumption on the size of $\phi$ is not restrictive in this case, if $\phi$ is a CNF formula.

## 3. Compilability Classes

We assume the reader is familiar with basic complexity classes, such as P, NP and (uniform) classes of the polynomial hierarchy (Stockmeyer, 1976; Garey & Johnson, 1979). Here we just briefly introduce non-uniform classes (Johnson, 1990). In the sequel, C, C', etc. denote arbitrary classes of the polynomial hierarchy.

We assume that the input instances of problems are strings built over an alphabet $\Sigma$. We denote with $\epsilon$ the empty string and assume that the alphabet $\Sigma$ contains a special symbol # to denote blanks. The *length* of a string $x \in \Sigma^*$ is denoted by $|x|$.

**Definition 1** *An advice $A$ is a function that takes an integer and returns a string.*

Advices are important in complexity theory because definitions and results are often based on special Turing machines that can determine the result of an oracle "for free", that is, in constant time.

**Definition 2** *An* advice-taking Turing machine *is a Turing machine enhanced with the possibility to determine $A(|x|)$ in constant time, where $x$ is the input string.*

Of course, the fact that $A(|x|)$ can be determined in constant time (while $A$ can be an intractable or even undecidable function) makes all definitions based on advice-taking Turing machine different from the same ones based on regular Turing machine. For example, an advice-taking Turing machine can calculate in polynomial time many functions that a regular Turing machine cannot (including some untractable ones).

Note that the advice is only a function of the *size* of the input, not of the input itself. Hence, advice-taking Turing machines are closely related to non-uniform families of circuits (Boppana & Sipser, 1990). Clearly, if the advice were allowed to access the whole instance, it would be able to determine the solution of any problem in constant time.

**Definition 3** *An advice-taking Turing machine uses* polynomial advice *if there exists a polynomial $p$ such that the advice oracle $A$ satisfies $|A(n)| \le p(n)$ for any nonnegative integers $n$.*

The non-uniform complexity classes are based on advice-taking Turing machines. In this paper we consider a simplified definition, based on classes of the polynomial hierarchy.

**Definition 4** *If* C *is a class of the polynomial hierarchy, then* C/poly *is the class of languages defined by Turing machines with the same time bounds as* C, *augmented by polynomial advice.*

Any class C/poly is also known as *non-uniform* C, where non-uniformity is due to the presence of the advice. Non-uniform and uniform complexity classes are related: Karp and Lipton (1980) proved that if NP $\subseteq$ P/poly then $\Pi_2^p = \Sigma_2^p = $ PH, i.e., the polynomial hierarchy collapses at the second level, while Yap (1983) generalized their results, in particular by showing that if NP $\subseteq$ coNP/poly then $\Pi_3^p = \Sigma_3^p = $ PH, i.e., the polynomial hierarchy collapses at the third level. An improvement of this results has been given by Köbler and Watanabe (1998): they proved that $\Pi_k^p \subseteq \Sigma_k^p$/poly implies that the polynomial hierarchy collapses to ZPP($\Sigma_{k+1}^p$). The collapse of the polynomial hierarchy is considered very unlikely by most researchers in structural complexity.

We now summarize some definitions and results proposed to formalize the compilability of problems (Cadoli et al., 1996b), adapting them to the context and terminology of PKR formalisms. We remark that it is not the aim of this paper to give a formalization of compilability of problems, or to analyze problems from this point of view. Rather, we show how to *use* the compilability classes as a technical tool for proving results on the relative efficiency of formalisms in representing knowledge in little space.

Several papers in the literature focus on the problem of reducing the complexity of problems via a preprocessing phase (Kautz & Selman, 1992; Kautz et al., 1995; Khardon & Roth, 1997). This motivates the introduction of a measure of complexity of problems assuming that such preprocessing is allowed. Following the intuition that a knowledge base is known well before questions are posed to it, we divide a reasoning problem into two parts: one part is *fixed* or *accessible off-line* (the knowledge base), and the second one is *varying*, or *accessible on-line* (the interpretation/formula). Compilability aims at capturing the *on-line complexity* of solving a problem composed of such inputs, i.e., complexity with respect to the first input when the first one can be preprocessed in an arbitrary way. In the next section we show the close connection between compilability and the space efficiency of PKR formalisms.

A function $f$ is called *poly-size* if there exists a polynomial $p$ such that for all strings $x$ it holds $|f(x)| \leq p(|x|)$. An exception to this definition is when $x$ represents a number: in this case, we impose $|f(x)| \leq p(x)$. As a result, we can say that the function $A$ used in advice-taking turing machine is a polysize function.

A function $g$ is called *poly-time* if there exists a polynomial $q$ such that for all $x$, $g(x)$ can be computed in time less than or equal to $q(|x|)$. These definitions easily extend to binary functions as usual.

We define a *language of pairs S* as a subset of $\Sigma^* \times \Sigma^*$. This is necessary to represent the two inputs to a PKR reasoning problem, i.e., the knowledge base (KB), and the formula or interpretation. As an example, the problem of Inference in Propositional Logic (PLI) is defined as follows.

PLI = $\{\langle x, y\rangle \mid x$ is a set of propositional formulae (the KB), $y$ is a formula, and $x \vdash y\}$

It is well known that PLI is coNP-complete, i.e., it is one of the "hardest" problems among those belonging to coNP. Our goal is to prove that PLI is the "hardest" theorem-proving problem among those in coNP that can be solved by preprocessing the first input in an arbitrary way, i.e., the KB. To this end, we introduce a new hierarchy of classes, the *non-uniform compilability classes*, denoted as $\Vdash\!\!\leadsto$C, where C is a generic uniform complexity class, such as P, NP, coNP, or $\Sigma_2^p$.

**Definition 5 ($\Vdash\!\!\leadsto$C classes)** *A language of pairs $S \subseteq \Sigma^* \times \Sigma^*$ belongs to $\Vdash\!\!\leadsto$C iff there exists a binary poly-size function $f$ and a language of pairs $S' \in$ C such that for all $\langle x, y \rangle \in S$ it holds:*

$$\langle f(x, |y|), y \rangle \in S' \text{ iff } \langle x, y \rangle \in S$$

Notice that the poly-size function $f$ takes as input both $x$ (the KB) and the size of $y$ (either the formula or the interpretation). This is done for technical reason, that is, such assumption allows obtaining results that are impossible to prove if the function $f$ only takes $x$ as input (Cadoli et al., 1996b). Such assuption is useful for proving negative results, that is, theorems of impossibility of compilation: indeed, if it is impossible to reduce the complexity of a problem using a function that takes both $x$ and $|y|$ as input, then such reduction is also impossible using a function taking $x$ only as its argument.

**Theorem 1 (Cadoli, Donini, Liberatore, & Schaerf, 1997, Theorem 6)** *Let $C$ be a class in the polynomial hierarchy and $S \subseteq \Sigma^* \times \Sigma^*$. A problem $S$ belongs to $\Vdash\!\!\leadsto$C if and only if there exists a poly-size function $f$ and a language of pairs $S'$, such that for all $\langle x, y \rangle \in \Sigma^* \times \Sigma^*$ it holds that:*

1. *for all $y$ such that $|y| \leq k$, $\langle f(x, k), y \rangle \in S'$ if and only if $\langle x, y \rangle \in S$;*

2. *$S' \in$ C.*

Clearly, any problem whose time complexity is in C is also in $\Vdash\!\!\leadsto$C: just take $f(x, |y|) = x$ and $S' = S$. What is interesting is that some problem in C may belong to $\Vdash\!\!\leadsto$C' with C' $\subset$ C, e.g.,, some problems in NP are in $\Vdash\!\!\leadsto$P. This is true for example for some problems in belief revision (Cadoli et al., 1999). In the rest of this paper, however, we mainly focus on "complete" problems, defined below. A pictorial representation of the class $\Vdash\!\!\leadsto$C is in Figure 1, where we assume that $S' \in C$.

For the problem PLI no method proving that it belongs to $\Vdash\!\!\leadsto$P is known. In order to show that it (probably) does not belong to $\Vdash\!\!\leadsto$P, we define a notion of reduction and completeness.

**Definition 6 (Non-uniform comp-reducibility)** *Given two problems $A$ and $B$, $A$ is non-uniformly comp-reducible to $B$ (denoted as $A \leq_{nu-comp} B$) iff there exist two poly-size binary functions $f_1$ and $f_2$, and a polynomial-time binary function $g$ such that for every pair $\langle x, y \rangle$ it holds that $\langle x, y \rangle \in A$ if and only if $\langle f_1(x, |y|), g(f_2(x, |y|), y) \rangle \in B$.*

The $\leq_{nu-comp}$ reductions can be represented as depicted in Figure 2. Such reductions satisfy all important properties of a reduction.

**Theorem 2 (Cadoli et al., 1996b, Theorem 5)** *The reductions $\leq_{nu-comp}$ satisfy transitivity and are compatible (Johnson, 1990) with the class $\Vdash\!\!\leadsto$C for every complexity class C.*

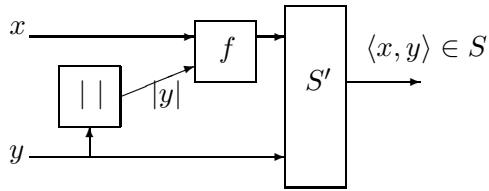

Figure 1: A representation of $\Vdash\!\!\rightsquigarrow$C.

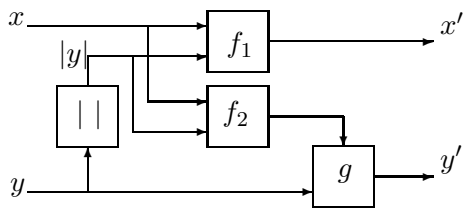

Figure 2: The nu-comp-C reductions.

Therefore, it is possible to define the notions of *hardness* and *completeness* for $\Vdash\!\!\rightsquigarrow$C for every complexity class C.

**Definition 7 ($\Vdash\!\!\rightsquigarrow$C-completeness)** *Let $S$ be a language of pairs and* C *a complexity class. $S$ is $\Vdash\!\!\rightsquigarrow$C-hard iff for all problems $A \in \Vdash\!\!\rightsquigarrow$C we have that $A \leq_{nu-comp} S$. Moreover, $S$ is $\Vdash\!\!\rightsquigarrow$C-complete if $S$ is in $\Vdash\!\!\rightsquigarrow$C and is $\Vdash\!\!\rightsquigarrow$C-hard.*

We now have the right complexity class to completely characterize the problem PLI. In fact PLI is $\Vdash\!\!\rightsquigarrow$coNP-complete (Cadoli et al., 1996b, Theorem 7). Furthermore, the hierarchy formed by the compilability classes is proper if and only if the polynomial hierarchy is proper (Cadoli et al., 1996b; Karp & Lipton, 1980; Yap, 1983) — a fact widely conjectured to be true.

Informally, we may say that $\Vdash\!\!\rightsquigarrow$NP-hard problems are "not compilable to P", as from the above considerations we know that if there exists a preprocessing of their fixed part that makes them on-line solvable in polynomial time, then the polynomial hierarchy collapses. The same holds for $\Vdash\!\!\rightsquigarrow$coNP-hard problems. In general, a problem which is $\Vdash\!\!\rightsquigarrow$C-complete for a class C can be regarded as the "toughest" problem in C, even after arbitrary preprocessing of the fixed part. On the other hand, a problem in $\Vdash\!\!\rightsquigarrow$C is a problem that, after preprocessing of the fixed part, becomes a problem in C (i.e., it is "compilable to C").

We close the section by giving another example of use of the compilability classes through the well-known formalism of Circumscription (McCarthy, 1980). Let $x$ be any propositional formula. The *minimal* models of $x$ are the truth assignments satisfying $x$ having as few positive values as possible (w.r.t. set containment). The problem we consider is: check whether a given model is a minimal model of a propositional formula. This problem, called Minimal Model checking (MMC), can be reformulated as the problem of model checking in Circumscription, which is known to be co-NP-complete (Cadoli, 1992).

If we consider the knowledge base $x$ as given off-line, and the truth assignment $y$ as given on-line, we obtain the following definition:

$$\text{MMC} = \{\langle x, y\rangle \mid y \text{ is a minimal model of } x \}$$

This problem can be shown to be $\Vert\leadsto$coNP-complete (Cadoli et al., 1996b, Theorem 13). Hence, it is very unlikely that it can be in $\Vert\leadsto$P; that is, it is very unlikely that there exists some off-line processing of the knowledge base, yielding (say) some data structure $x'$, such that given $y$, it can now checked in polynomial time whether $y$ is a minimal model of $x$. This, of course, unless $x'$ has exponential size. This observation applies also when $x'$ is a knowledge base in Propositional Logic, and led to the interpretation that Circumscription is more compact, or succint, than PL (Cadoli, Donini, & Schaerf, 1995; Gogic et al., 1995). Our framework allows to generalize these results for all PKR formalisms, as shown in the sequel.

## 4. Reductions among KR Formalisms

We now define the forms of reduction between PKR formalisms that we analyze in the following sections. A formula can always be represented as a string over an alphabet $\Sigma$, hence from now on we consider translations as functions transforming strings.

Let $F_1$ and $F_2$ be two PKR formalisms. There exists a *poly-size reduction* from $F_1$ to $F_2$, denoted as $f : F_1 \mapsto F_2$, if $f$ is a poly-size function such that for any given knowledge base $KB$ in $F_1$, $f(KB)$ is a knowledge base in $F_2$. Clearly, reductions should be restricted to produce a meaningful output. In particular, we now discuss reductions that preserve the models of the original theory.

The semantic approach by Gogic and collegues (1995) is that the models of the two knowledge bases must be exactly the same. In other words, if a knowledge base $KB$ of the formalism $F_1$ is translated into a knowledge base $KB'$ of the formalism $F_2$, then $M \models_{F_1} KB$ if and only if $M \models_{F_2} KB'$. This approach can be summarized by: a reduction between formalisms $F_1$ and $F_2$ is a way to translate knowledge bases of $F_1$ into knowledge bases of $F_2$, preserving their sets of models. While this semantics is intuitively grounded, it is very easy to show examples in which two formalisms that we consider equally space-efficient cannot be translated to each other. Let us consider for instance a variant of the propositional calculus in which the syntax is that formulas must be of the form $x_1 \wedge F$, where $F$ is a regular formula over the variables $x_2, \ldots$. Clearly, this formalism is able to represent knowledge in the same space than the propositional calculus (apart a polynomial factor). However, according to the definition, this formalism cannot be translated to propositional calculus: there is no knowledge base that is equivalent to $KB = \neg x_1$. Indeed, the only model of $KB$ is $\emptyset$, while any model of any consistent knowledge base of the modified propositional calculus contains $x_1$.

We propose a more general approach that can deal also with functions $f$ that change the language of the $KB$. To this end, we allow for a translation $g_{KB}$ from models of $KB$ to models of $f(KB)$. We stress that, to be as general as possible, the translation may depend on $KB$ — i.e., different knowledge bases may have different translations of their models. We want this translation easy to compute, since otherwise the computation of $g_{KB}$ could hide the complexity of reasoning in the formalism. However, observe that to this end, it is not sufficient to impose that $g_{KB}$ is computable in polynomial time. In fact, once $KB$ is fixed, its models could be trivially translated to models of $f(KB)$ in constant time, using a lookup table. This table would be exponentially large, though; and this is what we want to forbid. Hence, we impose that $g_{KB}$ is a *circuit* of polynomial-size wrt $KB$. We still use a functional notation $g_{KB}(M)$ to denote the result of applying a model $M$ to the circuit $g_{KB}$. A formal definition follows.

**Definition 8 (Model Preservation)** *A poly-size reduction $f : F_1 \mapsto F_2$ satisfies* model-preservation *if there exists a polynomial $p$ such that, for each knowledge base $KB$ in $F_1$ there exists a circuit $g_{KB}$ whose size is bounded by $p(|KB|)$, and such that for every interpretation $M$ of the variables of $KB$ it holds that $M \models_{F_1} KB$ iff $g_{KB}(M) \models_{F_2} f(KB)$.*

The rationale of model-preserving reduction is that the knowledge base $KB$ of the first formalism $F_1$ can be converted into a knowledge base $f(KB)$ in the second one $F_2$, and this reduction should be such that each model $M$ in $F_1$ can be easily translated into a model $g_{KB}(M)$ in $F_2$.

We require $g$ to depend on $KB$, because the transformation $f$, in general, could take the actual form of $KB$ into account. This happens in the following example of a model-preserving translation.

**Example 3** *We reduce a fragment of skeptical default logic (Kautz & Selman, 1991) to circumscription with varying letters, using the transformation introduced by Etherington (1987). Let $\langle D, W \rangle$ be a prerequisite-free normal (PFN) default theory, i.e., all defaults are of the form $\frac{:\gamma}{\gamma}$, where $\gamma$ is a generic formula. Let $Z$ be the set of letters occurring in $\langle D, W \rangle$. Define $P_D$ as the set of letters $\{a_\gamma | \frac{:\gamma}{\gamma} \in D\}$. The function $f$ can be defined in the following way: $f(\langle D, W \rangle) = CIRC(T; P_D; Z)$, where $T = W \cup \{a_\gamma \equiv \neg\gamma | a_\gamma \in P_D\}$, $P_D$ are the letters to be minimized, and $Z$ (the set of letters occurring in $\langle D, W \rangle$) are varying letters. We show that $f$ is a model-preserving poly-size reduction. In fact, given a set of PFN defaults $D$ let $g_D$ be a function such that for each interpretation $M$ for $Z$, $g_D(M) = M \cup \{a_\gamma \in P_D | M \models \neg\gamma\}$. Clearly, $f$ is poly-size, $g_D$ can be realized by a circuit whose size is polynomial in $|D|$, and $M$ is a model of at least one extension of $\langle D, W \rangle$ iff $g_D(M) \models CIRC(T; P_D; Z)$. The dependence of $g$ only on $D$ stresses the fact that, in this case, the circuit $g$ does not depend on the whole knowledge base $\langle D, W \rangle$, but just on $D$.*

Clearly, when models are preserved, theorems are preserved as well. A weaker form of reduction is the following one, where only theorems are preserved. Also in this case we allow theorems of $KB$ to be translated by a "simple" circuit $g_{KB}$ to theorems of $KB$.

**Definition 9 (Theorem Preservation)** *A poly-size reduction $f : F_1 \mapsto F_2$ satisfies* theorem-preservation *if there exists a polynomial $p$ such that, for each knowledge base $KB$ in $F_1$, there exists a circuit $g_{KB}$ whose size is bounded by $p(|KB|)$, and such that for every formula $\varphi$ on the variables of $KB$, it holds that $KB \vdash_{F_1} \varphi$ iff $f(KB) \vdash_{F_2} g_{KB}(\varphi)$.*

The theorem-preserving reduction has a property similar to that of the model-preserving reduction, when the knowledge bases are used to represent theorems rather than models. Namely, a knowledge base $KB$ is translated into another knowledge base $f(KB)$ which can be used to represent the same set of theorems. More precisely, we have that each theorem $\varphi$ of $KB$ is represented by a theorem $g_{KB}(\varphi)$ of $f(KB)$.

Winslett (1989) has shown an example of a reduction from updated knowledge bases to PL that is theorem-preserving but not model-preserving. Using Winslett's reduction, one could use the same machinery for propositional reasoning in the $KB$, both before and after the update (plus the reduction). Also the reduction shown in the previous Example 3 is theorem-preserving, this time $g$ being the identity circuit.

We remark that our definitions of reduction are more general than those proposed by Gogic and collegues (1995). In fact, these authors consider only a notion analogous to Definition 8. and only for the case when $g$ is the identity — i.e., models in the two formalisms should be identical. By allowing a simple translation $g$ between models Definition 8 covers more general forms of reductions preserving models, like the one of Example 3.

## 5. Comparing the Space Efficiency of PKR Formalisms

In this section we show how to use the compilability classes defined in Section 3 to compare the succinctness of PKR formalisms.

Let $F_1$ and $F_2$ be two formalisms representing sets of models. We prove that any knowledge base $\alpha$ in $F_1$ can be reduced, via a poly-size reduction, to a knowledge base $\beta$ in $F_2$ satisfying model-preservation if and only if the compilability class of the problem of *model checking* (first input: KB, second input: interpretation) in $F_2$ is higher than or equal to the compilability class of the problem of model checking in $F_1$.

Similarly, we prove that theorem-preserving poly-size reductions exist if and only if the compilability class of the problem of *inference* (first input: KB, second input: formula, cf. definition of the problem PLI) in $F_1$ is higher than or equal to the compilability class of the problem of inference in $F_2$.

In order to simplify the presentation and proof of the theorems we introduce some definitions.

**Definition 10 (Model hardness/completeness)** *Let $F$ be a PKR formalism and* C *be a complexity class. If the problem of model checking for $F$ belongs to the compilability class* $\|{\sim}$C, *where the model is the varying part of the instances, we say that $F$ is in* model-C. *Similarly, if model checking is* $\|{\sim}$C-complete (hard), *we say that $F$ is* model-C-*complete (hard).*

**Definition 11 (Theorem hardness/completeness)** *Let $F$ be a PKR formalism and* C *be a complexity class. If the problem of inference for the formalism $F$ belongs to the compilability class* $\|{\sim}$C, *whenever the formula is the varying part of the instance, we say that $F$ is in* thm-C. *Similarly, if inference is* $\|{\sim}$C-complete (hard), *we say that $F$ is* thm-C-*complete (hard).*

These definitions implicitly define two hierarchies, which parallel the polynomial hierarchy (Stockmeyer, 1976): the model hierarchy (model-P,model-NP,model-$\Sigma_2^p$,etc.) and the theorem hierarchy (thm-P,thm-NP,thm-$\Sigma_2^p$,etc.). The higher a formalism is in the model hierarchy, the more its efficiency in representing models is — and analogously for theorems. As an example (Cadoli et al., 1996, Thm. 6), we characterize model and theorem classes of Propositional Logic.

**Theorem 3** PL *is in* model-P *and it is* thm-coNP-*complete.*

We can now formally establish the connection between succinctness of representations and compilability classes. In the following theorems, the complexity classes C, $C_1$, $C_2$ belong to the polynomial hierarchy (Stockmeyer, 1976). In Theorems 5 and 7 we assume that the polynomial hierarchy does not collapse.

We start by showing that the existence of model-preserving reductions from a formalism to another one can be easily obtained if their levels in the model hierarchy satisfy a simple condition.

**Theorem 4** *Let $F_1$ and $F_2$ be two PKR formalisms. If $F_1$ is in* model-C *and $F_2$ is* model-C-*hard, then there exists a poly-size reduction $f : F_1 \mapsto F_2$ satisfying model preservation.*

*Proof.* Recall that since $F_1$ is in model-C, model checking in $F_1$ is in $\|\leadsto C$, and since $F_2$ is model-C-hard, model checking in $F_1$ is non-uniformly comp-reducible to model checking in $F_2$. That is, (adapting Def. 6) there exist two poly-size binary functions $f_1$ and $f_2$, and a polynomial-time binary function $g$ such that for every pair $\langle KB, M \rangle$ it holds that

$$M \models_{F_1} KB \text{ if and only if } g(f_2(KB, |M|), M) \models_{F_2} f_1(KB, |M|)$$

(note that $g$ is the poly-time function appearing in Def. 6, different from $g_{KB}$ which is the poly-size circuit appearing in Def. 8).

Now observe that $|M|$ can be computed from $KB$ by simply counting the letters appearing in $KB$; let $f_3$ be such a counting function, i.e., $|M| = f_3(KB)$. Clearly, $f_3$ is poly-size. Define the reduction $f$ as $f(KB) = f_1(KB, f_3(KB))$. Since poly-size functions are closed under composition, $f$ is poly-size. Now we show that $f$ is a model-preserving reduction. By Definition 8, we need to prove that there exists a polynomial $p$ such that for each knowledge base $KB$ in $F_1$, there exists a poly-size circuit $g_{KB}$ such that for every interpretation $M$ of the variables of $KB$ it holds that $M \models_{F_1} KB$ iff $g_{KB}(M) \models_{F_2} f(KB)$.

We proceed as follows: Given a $KB$ in $F_1$, we compute $z = f_2(KB, |M|) = f_2(KB, f_3(KB))$. Since $f_2$ and $f_3$ are poly-size, $z$ has size polynomial with respect to $|KB|$. Define the circuit $g_{KB}(M)$ as the one computing $g(z, M) = g(f_2(KB, f_3(KB)), M)$. Since $g$ is a poly-time function over both inputs, and $z$ is poly-size in $KB$, there exists a representation of $g(z, M)$ as a circuit $g_{KB}$ whose size is polynomial wrt $KB$. From this construction, $M \models_{F_1} KB$ iff $g_{KB}(M) \models_{F_2} f(KB)$. Hence, the thesis follows. $\square$

The following theorem, instead, gives a simple method to prove that there is no model-preserving reduction from one formalism to another one.

**Theorem 5** *Let $F_1$ and $F_2$ be two PKR formalisms. If the polynomial hierarchy does not collapse, $F_1$ is* model-$C_1$-*hard, $F_2$ is in* model-$C_2$, *and $C_2 \subset C_1$, then there is no poly-size reduction $: F_1 \mapsto F_2$ satisfying model preservation.*

*Proof.* We show that if such a reduction exists, then $C_1/poly \subseteq C_2/poly$ which implies that the polynomial hierarchy collapses at some level (Yap, 1983). Let $A$ be a complete problem for class $C_1$ — e.g., if $C_1$ is $\Sigma_3^p$ then $A$ may be validity of $\exists\forall\exists$-quantified boolean formulae (Stockmeyer, 1976). Define the problem $\epsilon A$ as follows.

$$\epsilon A = \{\langle x, y \rangle \mid x = \epsilon \text{ (the empty string) and } y \in A\}$$

We already proved (Cadoli et al., 1996b, Thm. 6) that $\epsilon A$ is $\|\!\leadsto\!C_1$-complete. Since model checking in $F_1$ is model-$C_1$-hard, $\epsilon A$ is non-uniformly comp-reducible to model checking in $F_1$. That is, (adapting Def. 6) there exist two poly-size binary functions $f_1$ and $f_2$, and a polynomial-time binary function $g$ such that for every pair $\langle \epsilon, y \rangle$, it holds $\langle \epsilon, y \rangle \in \epsilon A$ if and only if $g(f_2(\epsilon, |y|), y) \models_{F_1} f_1(\epsilon, |y|)$. Let $|y| = n$. Clearly, the knowledge base $f_1(\epsilon, |y|)$ depends only on $n$, i.e., there is exactly one knowledge base for each integer. Call it $KB_n$. Moreover, $f_2(\epsilon, |y|) = f_2(\epsilon, n)$ also depends on $n$ only: call it $O_n$ (for Oracle). Observe that both $KB_n$ and $O_n$ have polynomial size with respect to $n$.

If there exists a poly-size reduction $f : F_1 \mapsto F_2$ satisfying model preservation, then given the knowledge base $KB_n$ there exists a poly-size circuit $h_n$ such that $g(O_n, y) \models_{F_1} KB_n$ if and only if $h_n(g(O_n, y)) \models_{F_2} f(KB_n)$.

Therefore, the $\|\!\leadsto\!C_1$-complete problem $\epsilon A$ can be non-uniformly reduced to a problem in $\|\!\leadsto\!C_2$ as follows: Given $y$, from its size $|y| = n$ one obtains (with a preprocessing) $f(KB_n)$ and $O_n$. Then one checks whether the interpretation $h_n(g(O_n, y))$ (computable in polynomial time given $n$, $y$ and $O_n$) is a model in $F_2$ for $f(KB_n)$. From the fact that model checking in $F_2$ is in $\|\!\leadsto\!C_2$, we have that $\|\!\leadsto\!C_1 \subseteq \|\!\leadsto\!C_2$. We proved in a previous paper that such result implies that $C_1/\text{poly} \subseteq C_2/\text{poly}$ (Cadoli et al., 1996b, Thm. 9), which in turns implies that the polynomial hierarchy collapses (Yap, 1983). $\qquad \square$

The above theorems show that the hierarchy of classes model-C exactly characterizes the space efficiency of a formalism in representing sets of models. In fact, two formalisms at the same level in the model hierarchy can be reduced into each other via a poly-size reduction (Theorem 4), while there is no poly-size reduction from a formalism ($F_1$) higher up in the hierarchy into one ($F_2$) in a lower class (Theorem 5). In the latter case we say that $F_1$ is *more space-efficient* than $F_2$.

Analogous results (with similar proofs) hold for poly-size reductions preserving theorems. Namely, the next theorem shows how to infer the existence of theorem-preserving reductions, while the other one gives a way to prove that there is no theorem-preserving reduction from one formalism to another one.

**Theorem 6** *Let $F_1$ and $F_2$ be two PKR formalisms. If $F_1$ is in thm-C and $F_2$ is thm-C-hard, then there exists a poly-size reduction $f : F_1 \mapsto F_2$ satisfying theorem preservation.*

*Proof.* Recall that since $F_1$ is in thm-C, inference in $F_1$ is in $\|\!\leadsto\!C$, and since $F_2$ is thm-C-hard, inference in $F_1$ is non-uniformly comp-reducible to inference in $F_2$. That is, (adapting Def. 6) there exist two poly-size binary functions $f_1$ and $f_2$, and a polynomial-time binary function $g_1$ such that for every pair $\langle KB, \varphi \rangle$ it holds that

$$KB \vdash_{F_1} \varphi \text{ if and only if } f_1(KB, |\varphi|) \vdash_{F_2} g(f_2(KB, |\varphi|), \varphi)$$

(here we distinguish the poly-time function $g$ appearing in Def. 6 and the poly-size circuit $g_{KB}$ appearing in Def. 9).

Using Theorem 1 we can replace $|\varphi|$ with an upper bound in the above formula. From Assumption 1, we know that the size of $\varphi$ is less than or equal to the size of $KB$; therefore we replace $|\varphi|$ with $|KB|$. The above formula now becomes

$$KB \vdash_{F_1} \varphi \text{ if and only if } f_1(KB, |KB|) \vdash_{F_2} g(f_2(KB, |KB|), \varphi)$$

Define the reduction $f$ as $f(KB) = f_1(KB, f_3(KB))$, where $f_3$ is the poly-size function that computes the size of its input. Since poly-size functions are closed under composition, $f$ is poly-size.

Now, we show that $f$ is a theorem-preserving reduction, i.e., $f$ satisfies Def. 9. This amounts to proving that for each knowledge base $KB$ in $F_1$ there exists a circuit $g_{KB}$, whose size is ponynomial wrt $KB$, such that for every formula $\varphi$ on the variables of $KB$ it holds that $KB \vdash_{F_1} \varphi$ iff $f(KB) \vdash_{F_2} g_{KB}(\varphi)$.

We proceed as in the proof of Theorem 4: Given a $KB$ in $F_1$, let $z = f_2(KB, f_3(KB))$. Since $f_2$ and $f_3$ are poly-size, $z$ has polynomial size with respect to $|KB|$. Define $g_{KB}(\varphi) = g(z, \varphi) = g(f_2(KB, f_3(KB)), \varphi)$. Clearly, $g_{KB}$ can be represented by a circuit of polynomial size wrt $KB$. From this construction, $KB \vdash_{F_1} \varphi$ iff $f(KB) \vdash_{F_2} g_{KB}(\varphi)$. Hence, the claim follows. $\qquad\blacksquare$

**Theorem 7** *Let $F_1$ and $F_2$ be two PKR formalisms. If the polynomial hierarchy does not collapse, $F_1$ is thm-$C_1$-hard, $F_2$ is in thm-$C_2$, and $C_2 \subset C_1$, then there is no poly-size reduction $f : F_1 \mapsto F_2$ satisfying theorem preservation.*

*Proof.* We show that if such a reduction exists, then $C_1/\text{poly} \subseteq C_2/\text{poly}$ and the polynomial hierarchy collapses at some level (Yap, 1983). Let $A$ be a complete problem for class $C_1$. Define the problem $\epsilon A$ as in the proof of Theorem 5: this problem is $\Vdash\!\!\sim C_1$-complete (Cadoli et al., 1996b, Thm. 6). Since inference in $F_1$ is thm-$C_1$-hard, $\epsilon A$ is non-uniformly comp-reducible to inference in $F_1$. That is, (adapting Def. 6) there exist two poly-size binary functions $f_1$ and $f_2$, and a polynomial-time binary function $g$ such that for every pair $\langle \epsilon, y \rangle$, $\langle \epsilon, y \rangle \in \epsilon A$ if and only if $f_1(\epsilon, |y|) \vdash_{F_1} g(f_2(\epsilon, |y|), y)$. Let $|y| = n$. Clearly, the knowledge base $f_1(\epsilon, |y|)$ depends just on $n$, i.e., there is one knowledge base for each integer. Call it $KB_n$. Moreover, also $f_2(\epsilon, |y|) = f_2(\epsilon, n)$ depends just on $n$: call it $O_n$ (for Oracle). Observe that both $KB_n$ and $O_n$ have polynomial size with respect to $n$.

If there exists a poly-size reduction $f : F_1 \mapsto F_2$ satisfying theorem preservation, then given the knowledge base $KB_n$ there exists a poly-time function $h_n$ such that $KB_n \vdash_{F_1} g(O_n, y)$ if and only if $f(KB_n) \vdash_{F_2} h_n(g(O_n, y))$.

Therefore, the $\Vdash\!\!\sim C_1$-complete problem $\epsilon A$ can be non-uniformly reduced to a problem in $\Vdash\!\!\sim C_2$ as follows: Given $y$, from its size $|y| = n$ one obtains (with an arbitrary preprocessing) $f(KB_n)$ and $O_n$. Then one checks whether the formula $h_n(g(O_n, y))$ (computable in poly-time given $y$ and $O_n$) is a theorem in $F_2$ of $f(KB_n)$. From the fact that inference in $F_2$ is in $\Vdash\!\!\sim C_2$, we have that $\Vdash\!\!\sim C_1 \subseteq \Vdash\!\!\sim C_2$. It follows that $C_1/\text{poly} \subseteq C_2/\text{poly}$ (Cadoli et al., 1996b, Thm. 9), which implies that the polynomial hierarchy collapses (Yap, 1983). $\qquad\blacksquare$

Theorems 4-7 show that compilability classes characterize very precisely the relative capability of PKR formalisms to represent sets of models or sets of theorems. For example, as a consequence of Theorems 3 and 7 there is no poly-size reduction from PL to the syntactic restriction of PL allowing only Horn clauses that preserves the theorems, unless the polynomial hierarchy collapses. Kautz and Selman (1992) proved non-existence of such a reduction for a problem strictly related to PLI using a specific proof.

## 6. Applications

This section is devoted to the application of the theorems presented in the previous section. Using Theorems 4-7 and results previously known from the literature, we are able to asses model- and theorem-compactness of some PKR formalisms.

We assume that definitions of Propositional Logic, default logic (Reiter, 1980), and circumscription (McCarthy, 1980) are known. Definitions of WIDTIO, SBR, GCWA, and stable model semantics are in the appropriate subsections.

In the following proofs we refer to the problem $\exists\forall 3$QBF, that is, the problem of verifying whether a quantified Boolean formula $\exists X \forall Y.\neg F$ is valid, where $X$ and $Y$ are disjoint sets of variables, and $F$ is a set of clauses on the alphabet $X \cup Y$, each composed of three literals. As an example, a simple formula belonging to this class is: $\exists x_1, x_2 \forall y_1, y_2 \quad \neg((x_1 \lor y_2) \land (\neg x_1 \lor \neg x_2 \lor \neg y_1) \land (\neg y_1 \lor \neg x_2 \lor \neg y_2) \land (\neg x_1 \lor \neg x_2))$.

The problem of deciding validity of a $\exists\forall 3$QBF is complete for the class $\Sigma_2^p$. As a consequence, the corresponding problem $*\exists\forall 3$QBF, that is deciding whether an input composed of any string ($*$) as the fixed part and a quantified Boolean formula $\exists X \forall Y.\neg F$ as the varying one, is complete for the class $\|\leadsto\Sigma_2^p$ (Liberatore, 1998). Notice that in most of the hardness proofs we show in the sequel we use problems without any meaningful fixed part.

### 6.1 Stable Model Semantics

Stable model semantics (**SM**) was introduced by Gelfond and Lifschitz (1988) as a tool to provide a semantics for logic programs with negation. their original proposal is now one of the standard semantics for logic programs. We now recall the definition of propositional stable model.

Let $P$ be a propositional, general logic program. Let $M$ be a subset (i.e., an interpretation) of the atoms of $P$. Let $P^M$ be the program obtained from $P$ in the following way: if a clause $C$ of $P$ contains in its body a negated atom $\neg A$ such that $A \in M$ then $C$ is deleted; if a body of a clause contains a negated atom $\neg A$ such that $A \notin M$ then $\neg A$ is deleted from the body of the clause. If $M$ is a least Herbrand model of $P^M$ then $M$ is a stable model of $P$.

For the formalism sm, we consider the program $P$ as the knowledge base. We write $P \models_{\text{SM}} Q$ to denote that query $Q$ is implied by a logic program $P$ under Stable Model semantics.

In order to prove our result, we need to define the kernel of a graph.

**Definition 12 (Kernel)** *Let $G = (V, E)$ be a graph. A* kernel *of $G$ is a set $K \subseteq V$ such that, denoting $H = V - K$, it holds:*

  1. *$H$ is a vertex cover of $G$*

  2. *for all $j \in H$, there exists an $i \in K$ such that $(i, j) \in E$.*

We can now state the theorem on the compilability class of inference in the stable model semantics, and the corresponding theorem compactness class.

**Theorem 8** *The problem of inference for the Stable Model semantics is $\|\leadsto$coNP-complete, thus Stable Model Semantics is* thm-coNP *complete.*

*Proof.* Membership in the class follows from the fact that the problem is coNP-complete (Marek & Truszczyński, 1991). For the hardness, we adapt the proof of Marek and Truszczyński (1991) showing that deciding whether a query is true in all stable models is coNP-hard.

Let $\epsilon$KERNEL be the language $\{\epsilon, G\}$ such that $G$ is a graph with at least one kernel. Let $|G| = n$, and observe that $G$ cannot have more vertices than its size $n$.

We show that for each $n$, there exists a logic program $P_n$ such that for every graph $G$ with at most $n$ vertices, there exists a query $Q_G$ such that $G$ has a kernel iff $P_n \not\models_{\text{SM}} Q_G$.

Let the alphabet of $P_n$ be composed by the following $2n^2 + n$ propositional letters: $\{a_i | i \in \{1..n\}\} \cup \{r_{ij}, s_{ij} | i, j \in \{1..n\}\}$.

The program $P_n$ is defined as:

$$\left. \begin{array}{rcl} a_j & :- & \neg a_i, r_{ij} \\ s_{ij} & :- & \neg r_{ij} \\ r_{ij} & :- & \neg s_{ij} \end{array} \right\} \text{for } i, j \in \{1..n\}$$

Given a graph $G = (V, E)$, the query $Q_G$ is defined as

$$Q_G = (\bigvee_{(i,j) \in E} \neg r_{ij}) \vee (\bigvee_{(i,j) \notin E} r_{ij})$$

The reduction from $\epsilon$KERNEL to SM is defined as: $f_1(x, n) = P_n$, i.e., $f_1$ depends only on its second argument, $f_2(x, n) = \epsilon$, i.e., $f_2$ is a constant function, and $g = Q_y$, i.e., given a graph $G$, the circuit $g$ computes the query $Q_G$.

As a result, this is a $\|\rightsquigarrow$ reduction. We now show that this reduction is correct, i.e., $\langle \epsilon, G \rangle \in \epsilon$KERNEL ($G$ has a kernel) iff $P_n \not\models_{SM} Q_G$.

*If-part.* Suppose $P_n \not\models_{SM} Q_G$. Then, there exists a stable model $M$ of $P_n$ such that $M \models \neg Q_G$. Observe that $\neg Q_G$ is equivalent to the conjunction of all $r_{ij}$ such that $(i, j) \in E$, and all $\neg r_{ij}$ such that $(i, j) \notin E$. Simplifying $P_n$ with $\neg Q_G$ we obtain the clauses:

$$a_j \ :- \neg a_i, \text{ for } (i, j) \in E \qquad (1)$$

Observe that $M$ contains all $s_{ij}$ such that $(i, j) \notin E$, and in order to be stable, — i.e., to support atoms $r_{ij}$ such that $(i, j) \in E$ — $M$ contains no atom $s_{ij}$ such that $(i, j) \in E$.

Let $H = \{j | a_j \in M\}$, $K = \{i | a_i \notin M\}$. Now $H$ is a vertex cover of G, since for each edge $(i, j) \in E$, $M$ should satisfy the corresponding clause (1) $a_j \ :- \neg a_i$, hence either $a_i \in M$, or $a_j \in M$. Moreover, for each $j$ in $H$, the atom $a_j$ is in $M$, and since $M$ is a stable model, there exists a clause $a_j \ :- \neg a_i$ such that $a_i \notin M$, that is, $i \in K$. Therefore, $K$ is a kernel of $G$.

*Only-if part.* Suppose $G = (V, E)$ has a kernel $K$, and let $H = V - K$. Let $M$ be the interpretation

$$M = \{r_{ij} | (i, j) \in E\} \cup \{s_{ij} | (i, j) \notin E\} \cup \{a_j | j \in H\}$$

Obviously, $M \not\models Q_G$. We now show that $M$ is a stable model of $P_n$, i.e., $M$ is a least Herbrand model of $P_n^M$. In fact, $P_n^M$ contains the following clauses:

$$s_{ij} \qquad\qquad \text{for } (i, j) \notin E \qquad (2)$$

$$r_{ij} \qquad\qquad \text{for } (i, j) \in E \qquad (3)$$

$$a_j \qquad :- r_{ij} \quad \text{for } i \in K \qquad (4)$$

Clauses in the last line are obtained from clauses in $P_n$ of the form $a_j :- \neg a_i, r_{ij}$, where the clauses such that $i \in H$ (hence $a_i \in M$) are deleted, while in the other clauses the negated atom $\neg a_i$ is deleted, since $i \in K$, hence $a_i \notin M$. Now for each $a_j \in M$, the vertex $j$ is in $H$, hence there is an edge $(i,j) \in E$, and $i \in K$. Hence clauses (4) and (3) are in $P_n^M$, hence in the least Herbrand model of $P_n^M$ there are exactly all $a_j$ such that $j \in H$. ∎

### 6.2 Minimal Model Reasoning

One of the most successful form of non-monotonic reasoning is based on the selection of minimal models. Among the various formalisms based on minimal model semantics we consider here Circumscription (McCarthy, 1980) and the Generalized Closed World Assumption (GCWA) (Minker, 1982), which is a formalism to represent knowledge in a closed world.

We assume that the reader is familiar with Circumscription, we briefly present the definition of GCWA. The model semantics for GCWA is defined as ($a$ is a letter):

$M \models_{GCWA} KB$ iff $M \models KB \cup \{\neg a \mid$ for any positive clause $\gamma$, if $KB \not\vdash \gamma$ then $KB \not\vdash \gamma \vee a\}$

We can now present the results for these two formalisms.

**Theorem 9** *The problem of model checking for Circumscription is* $\|{\rightsquigarrow}$coNP-*complete, thus Circumscription is* model-coNP-*complete.*

This result is a trivial corollary of a theorem already proved (Cadoli et al., 1997, Theorem 6). In fact, that proof implicitly shows that model checking for circumscription is $\|{\rightsquigarrow}$coNP-complete.

**Theorem 10** *The problem of model checking for GCWA is in* $\|{\rightsquigarrow}$P, *thus GCWA is in* model-P.

*Proof.* As already pointed out (Cadoli et al., 1997), it is possible to rewrite $GCWA(T)$ into a propositional formula $F$ such that, for any given model $M$, $M \models GCWA(T)$ if and only if $M \models F$. Moreover, the size of $F$ is polynomially bounded by the size of $T$. As a consequence, the model compactness for GCWA is in the same class of PL. By Theorem 3 the thesis follows. ∎

**Theorem 11** *The problem of inference for Circumscription is* $\|{\rightsquigarrow}\Pi_2^p$-*complete, thus Circumscription is* thm-$\Pi_2^p$-*complete.*

This result is a trivial corollary of a theorem published in a previous paper (Cadoli et al., 1997, Theorem 7) which implicitly shows that inference for circumscription is $\|{\rightsquigarrow}\Pi_2^p$-complete.

**Theorem 12** *The problem of inference for GCWA is* $\|{\rightsquigarrow}$coNP-*complete, thus GCWA is* thm-coNP-*complete.*

*Proof.* As already pointed out in the proof of Theorem 10, it is possible to rewrite $GCWA(T)$ into a formula $F$ that is equivalent to it. As a consequence, a formula $\alpha$ is a theorem of $GCWA(T)$ if and only if it is a theorem of $F$. Thus, GCWA has at most the theorem compexity of PL. Since GCWA is a generalization of PL, it follows that GCWA is in the same theorem compactness class of PL. Hence, GCWA is thm-coNP-complete. ∎

### 6.3 Default Logic

In this subsection we present the results for default logic, in its two variants (credulous and skeptical). For more details on these two main variants of default logic, we refer the reader to the paper by Kautz and Selman (1991). Notice that model-compactness is only applicable to skeptical default logic.

**Theorem 13** *The problem of model checking for skeptical default logic is $\Vdash\!\!\rightsquigarrow\Sigma_2^p$ complete, thus skeptical default logic is* model-$\Sigma_2^p$ *complete.*

*Proof.* The proof of membership is straightforward: since model checking for skeptical default logic is in $\Sigma_2^p$ (Liberatore & Schaerf, 1998), it follows that it is also in $\Vdash\!\!\rightsquigarrow\Sigma_2^p$.

The proof of $\Vdash\!\!\rightsquigarrow\Sigma_2^p$-hardness is similar to the proof of $\Sigma_2^p$-hardness (Liberatore & Schaerf, 1998). The reduction is from the problem $*\exists\forall 3$QBF. Let $\langle\alpha,\beta\rangle$ be an instance of $*\exists\forall 3$QBF, where $\beta = \exists X \forall Y. \neg F$ represents a valid $\exists\forall 3$QBF formula, and $\alpha$ is any string.

Let $n$ be the size of the formula $F$. This implies that the variables in the formula are at most $n$. Let $\Gamma = \{\gamma_1, \ldots, \gamma_k\}$ be the set of all the clauses of three literals over this alphabet. The number of clauses of three literals over an alphabet of $n$ variables is less than $O(n^3)$, thus bounded by a polynomial in $n$.

We prove that $\exists X \forall Y. \neg F$ is valid if and only if $M$ is a model of some extension of $\langle W, D \rangle$, where

$$W = \emptyset$$
$$D = \bigcup_{\gamma_i \in \Gamma} \left\{ \frac{:c_i}{c_i}, \frac{:\neg c_i}{\neg c_i} \right\} \cup \bigcup_{x_i \in X} \left\{ \frac{:w \wedge (w \rightarrow x_i)}{w \rightarrow x_i}, \frac{:w \wedge (w \rightarrow \neg x_i)}{w \rightarrow \neg x_i} \right\} \cup \left\{ \frac{:w \wedge \bigwedge_{\gamma_i \in \Gamma} c_i \rightarrow \gamma_i}{w} \right\}$$
$$M = \{c_i \mid \gamma_i \in F\}$$

The set $\{c_i \mid 1 \leq i \leq k\}$ is a set of new variables, one-to-one with the elements of $\Gamma$. Note that $W$ and $D$ only depends on the size $n$ of $F$, while $M$ depends on $F$. As a result, this is a $\leq_{nu-comp}$ reduction.

We now prove that the formula is valid if and only if $M$ is a model of some extension of the default theory $\langle W, D \rangle$. This is similar to an already published proof (Liberatore & Schaerf, 1998). Consider an evaluation $C_1$ of the variables $\{c_i\}$ and an evaluation $X_1$ of the variables $X$. Let $D'$ be the following set of defaults.

$$D' = \bigcup_{c_i \in C_1} \left\{ \frac{:c_i}{c_i} \right\} \bigcup_{c_i \notin C_1} \left\{ \frac{:\neg c_i}{\neg c_i} \right\} \cup \bigcup_{x_i \in X_1} \left\{ \frac{:w \wedge (w \rightarrow x_i)}{w \rightarrow x_i} \right\} \bigcup_{x_i \in \overline{X_1}} \left\{ \frac{:w \wedge (w \rightarrow \neg x_i)}{w \rightarrow \neg x_i} \right\}$$

This set of defaults has been chosen so that the set $R$ of its consequences corresponds to the sets $C_1$ and $X_1$. Namely, we have:

$$c_i \in C_1 \quad \text{iff} \quad R \models c_i$$
$$c_i \notin C_1 \quad \text{iff} \quad R \models \neg c_i$$
$$x_i \in X_1 \quad \text{iff} \quad R \models w \rightarrow x_i$$
$$x_i \notin X_1 \quad \text{iff} \quad R \models w \rightarrow \neg x_i$$

Now, we prove that the consequences of this set of defaults are an extension of the default theory if and only if the QBF formula is valid. Since all defaults are semi-normal, we have to prove that:

1. the set of consequences of $D'$ is consistent; and

2. no other default is applicable, that is, there is no other default whose precondition is consistent with $R$.

Consistency of $R$ follows by construction: assigning $c_i$ to true for each $c_i \in C_1$, etc., we obtain a model of $R$.

We have then to prove that no other default is applicable. If $c_i \in C_1$, the default $\frac{: \neg c_i}{\neg c_i}$ is not applicable, and vice versa, if $\neg c_i \in C_1$, then $\frac{: c_i}{c_i}$ is not applicable. Moreover, none of the defaults $\frac{:w \wedge (w \to x_i)}{w \to x_i}$, is applicable if $x_i \notin X_1$, because in this case $w \to \neg x_i \in R$, thus $\neg w$ would follow (while $w$ is a justification of the default). A similar statement holds for $\frac{:w \wedge (w \to \neg x_i)}{w \to \neg x_i}$ if $x_i \in X_1$.

As a result, the only applicable default may be the last one, $\frac{:w \wedge \bigwedge_{\gamma_i \in \Gamma} c_i \to \gamma_i}{w}$ (recall that F is negated). This default is applicable if and only if, for the given evaluation of the $c_i$'s and $x_i$'s, the set of clauses is satisfiable. This amount to say: "there is an extension in which the last default is not applicable if and only if the QBF formula is valid". Now, if the last default is applicable, then $M$ is not a model of the extension because $w$ is the consequence of the last default while $w \not\models M$. The converse also holds: if the last default is not applicable then $M$ is a model of the default theory.

As a result, the QBF is valid if and only if $M$ is a model of the given default theory. $\square$

**Theorem 14** *The inference problem for skeptical default logic is* $\Vdash\!\sim\!\Pi_2^p$ *complete, thus skeptical default logic is* thm-$\Pi_2^p$ *complete.*

*Proof.* Since inference in skeptical default logic is in $\Pi_2^p$, it is also in $\Vdash\!\sim\!\Pi_2^p$. $\Vdash\!\sim\!\Pi_2^p$-hardness comes from a simple reduction from circumscription. Indeed, the circumscription of a formula $T$ is equivalent to the conjunction of the extensions of the default theory $\langle T, D \rangle$, where (Etherington, 1987):

$$D = \bigcup \left\{ \frac{: \neg x_i}{\neg x_i} \right\}$$

As a result, $CIRC(T) \models Q$ if and only if $Q$ is implied by $\langle T, D \rangle$ under skeptical semantics. Since $\langle T, D \rangle$ only depends on $T$ (and not on $Q$) this is a $\leq_{nu-comp}$ reduction. Since inference for circumscription is $\Vdash\!\sim\!\Pi_2^p$-complete (see Theorem 11), it follows that skeptical default logic is $\Vdash\!\sim\!\Pi_2^p$-hard. $\blacksquare$

**Theorem 15** *The inference problem for credulous default logic is* $\Vdash\!\sim\!\Sigma_2^p$ *complete, thus credulous default logic is* thm-$\Sigma_2^p$ *complete.*

*Proof.* The proof is very similar to the proof for model checking of skeptical default logic. Indeed, both problems are $\|\!\sim\Sigma_2^p$ complete. Since the problem is in $\Sigma_2^p$, as proved by Gottlob (1992), it is also in $\|\!\sim\Sigma_2^p$. Thus, what we have to prove is that is hard for that class.

We prove that the $*\exists\forall$3QBF problem can be reduced to the problem of verifying whether a formula is implied by some extensions of a default theory (that is, inference in credulous default logic).

Namely, a formula $\forall X\exists Y.\neg F$ is valid if and only if $Q$ is derived by some extension of the default theory $\langle D, W\rangle$, where $W$ and $D$ are defined as follows ($\Gamma$ is the set of all the clauses of three literals over the alphabet of $F$, and $C$ is a set of new variables, one-to-one with $\Gamma$).

$$
\begin{aligned}
W &= \emptyset \\
D &= \bigcup_{c_i \in C} \left\{ \frac{:c_i}{c_i}, \frac{:\neg c_i}{\neg c_i} \right\} \cup \bigcup_{x_i \in X} \left\{ \frac{:x_i}{x_i}, \frac{:\neg x_i}{\neg x_i} \right\} \cup \left\{ \frac{\neg(\bigwedge_{c_i \in C} c_i \rightarrow \gamma_i):}{w} \right\} \\
Q &= \bigwedge_{\gamma_i \in F} c_i \wedge \bigwedge_{\gamma_i \notin F} \neg c_i \wedge w
\end{aligned}
$$

Informally, the proof goes as follows: for each truth evaluation of the variables in $C$ and $X$ there is a set of defaults which are both justified and consistent. A simple necessary and sufficient condition for the consequences of this set of defaults to be an extension is the following. If, in this evaluation, the formula

$$
\neg \bigwedge_{c_i = \text{true}} \gamma_i
$$

is valid, then the last default is applicable, thus the extension also contains $w$. The converse also holds: if the formula is not valid in the evaluation, then the variable $w$ is not in the extension.

As a result, there exists an extension in which $Q$ holds if and only if there exists an extension in which each $c_i$ is true if and only if $\gamma_i \in F$, and such that $w$ also holds. When the variables $c_i$ have the given value, the above formula is equivalent to $\neg F$. As a result, such an extension exists if and only if there exists a truth evaluation of the variables $X$ in which $\neg F$ is valid. □

## 6.4 Belief Revision

Many formalisms for belief revision have been proposed in the literature, here we focus on two of them: WIDTIO (When In Doubt Throw it Out) and SBR (Skeptical Belief Revision). Let $K$ be a set of propositional formulae, representing an agent's knowledge about the world. When a new formula $A$ is added to $K$, the problem of the possible inconsistency between $K$ and $A$ arises. The first step is to define the set of sets of formulae $W(K, A)$ in the following way:

$$
W(K, A) = \{ K' \mid K' \text{ is a maximal consistent subset of } K \cup \{A\} \text{ containing } A \}
$$

Any set of formulae $K' \in W(K, A)$ is a maximal choice of formulae in $K$ that are consistent with $A$ and, therefore, we may retain when incorporating $A$. The definition of this set leads to two different revision operators: SBR and WIDTIO.

**SBR** Skeptical Belief Revision (Fagin, Ullman, & Vardi, 1983; Ginsberg, 1986). The revised theory is defined as a set of theories: $K * A \doteq \{K' \mid K' \in W(K, A)\}$. Inference in the revised theory is defined as inference in each of the theories:

$$K * A \vdash_{SBR} Q \quad \text{iff} \quad \text{for all } K' \in W(K, A) \text{ , we have that } K' \vdash Q$$

The model semantics is defined as:

$$M \models_{SBR} K * A \quad \text{iff} \quad \text{there exists a } K' \in W(K, A) \text{ such that } M \models K'$$

**WIDTIO** When In Doubt Throw It Out (Winslett, 1990). A simpler (but somewhat drastical) approach is the so-called WIDTIO, where we retain only the formulae of $K$ that belong to all sets of $W(K, A)$. Thus, inference is defined as:

$$K * A \vdash_{WIDTIO} Q \quad \text{iff} \quad \bigcap W(K, A) \vdash Q$$

The model semantics of this formalism is defined as:

$$M \models_{WIDTIO} K * A \quad \text{iff} \quad M \models \bigcap W(K, A)$$

The results on model compactness have been shown by Liberatore and Schaerf (2000). Here we recall them.

**Theorem 16 (Liberatore & Schaerf, 2000, Theorem 11)** *The problem of model checking for WIDTIO is in* $\|\rightsquigarrow$P*, thus WIDTIO is in* model-P.

**Theorem 17 (Liberatore & Schaerf, 2000, Theorem 5)** *The problem of model checking for Skeptical Belief Revision is* $\|\rightsquigarrow$coNP*-complete, thus Skeptical Belief Revision is* model-coNP*-complete.*

The results on theorem compactness are quite simple and we provide here the proofs.

**Theorem 18** *The problem of inference for WIDTIO is* $\|\rightsquigarrow$coNP*-complete, thus WIDTIO is* thm-coNP*-complete.*

*Proof.* Membership in the class thm-coNP immediately follows from the definition. In fact, we can rewrite $K * A$ into a propositional formula by computing the set $W(K, A)$ and then constructing their intersection. By construction their intersection has size less than or equal to the size of $K \cup A$. As a consequence, after preprocessing, deciding whether a formula $Q$ follows from $K * A$ is a problem in coNP. Hardness follows from the obvious fact that PL can be reduced to WIDTIO and PL is thm-coNP-complete (see Theorem 3). □

**Theorem 19** *The problem of inference for Skeptical Belief Revision is* $\|\rightsquigarrow\Pi_2^p$*-complete, thus Skeptical Belief Revision is* thm-$\Pi_2^p$*-complete.*

| | Time Complexity | Space Efficiency |
|---|---|---|
| Propositional Logic | P<br>— | model-P<br>— |
| WIDTIO | $\Sigma_2^p$-complete<br>(Liberatore & Schaerf, 1996) | model-P<br>Th. 16 |
| Skeptical Belief Revision | coNP-complete<br>(Liberatore & Schaerf, 1996) | model-coNP-complete<br>Th. 17 |
| Circumscription | coNP-complete<br>(Cadoli, 1992) | model-coNP-complete<br>Th. 9 |
| GCWA | coNP-hard,<br>in $\Delta_2^p[\log n]$<br>(Eiter & Gottlob, 1993) | model-P<br>Th. 10 |
| Skeptical Default Reasoning | $\Sigma_2^p$-complete<br>(Liberatore & Schaerf, 1998) | model-$\Sigma_2^p$-complete<br>Th. 13 |
| Credulous Default Reasoning | N/A | N/A |
| Stable Model Semantics | P<br>— | model-P<br>— |

Table 1: Complexity of model checking and Space Efficiency of Model Representations

*Proof.* Membership follows from the complexity results of Eiter and Gottlob (1992), where they show that deciding whether $K * A \vdash_{SBR} Q$ is a $\Pi_2^p$-complete problem. Hardness follows easily from Theorem 17. In fact, $M \models_{SBR} K * A$ iff $K * A \nvdash_{SBR} \neg form(M)$, where $form(M)$ is the formula that represents the model $M$. As a consequence, model checking can be reduced to the complement of inference. Thus inference is $\|\!\sim\!\Pi_2^p$-complete. $\qquad\Box$

## 6.5 Discussion

Tables 1 and 2 summarize the results on space efficiency of PKR formalisms and where they were proved (a dash "—" denotes a folklore result).

First of all, notice that space efficiency is not always related to time complexity. As an example, we compare in detail WIDTIO and circumscription. From the table it follows that model checking is harder for WIDTIO than for circumscription, and that inference has the same complexity in both cases. Nevertheless, since circumscription is thm-$\Sigma_2^p$-complete and WIDTIO is thm-coNP-complete (and thus in thm-$\Sigma_2^p$), there exists a poly-size reduction from WIDTIO to circumscription satisfying theorem preservation. The converse does not hold: since circumscription is thm-$\Sigma_2^p$-complete and WIDTIO is thm-coNP, unless the Polynomial Hierarchy does not collapse there is no theorem-preserving poly-size reduction from the former formalism to the latter. Hence, circumscription is a more compact formalism than WIDTIO to represent theorems. Analogous considerations can be done for models. Intuitively, this is due to the fact that for WIDTIO both model checking and inference require a lot of work on the revised knowledge base alone—computing the intersection of all elements of $W(K, A)$. Once this is done, one is left with model checking and inference in PL. Hence, WIDTIO has the same space efficiency as PL, which is below circumscription.

| | Time Complexity | Space Efficiency |
|---|---|---|
| Propositional Logic | coNP-complete (Cook, 1971) | thm-coNP-complete (Cadoli et al., 1996) |
| WIDTIO | $\Pi_2^p$-complete (Eiter & Gottlob, 1992) & (Nebel, 1998) | thm-coNP-complete Th. 18 |
| Skeptical Belief Revision | $\Pi_2^p$-complete (Eiter & Gottlob, 1992) | thm-$\Pi_2^p$-complete Th. 19 |
| Circumscription | $\Pi_2^p$-complete (Eiter & Gottlob, 1993) | thm-$\Pi_2^p$-complete Th. 11 |
| GCWA | $\Pi_2^p$-complete (Eiter & Gottlob, 1993) & (Nebel, 1998) | thm-coNP-complete Th. 12 |
| Skeptical Default Reasoning | $\Pi_2^p$-complete (Gottlob, 1992) | thm-$\Pi_2^p$-complete Th. 14 |
| Credulous Default Reasoning | $\Sigma_2^p$-complete (Gottlob, 1992) | thm-$\Sigma_2^p$-complete Th. 15 |
| Stable Model Semantics | coNP-complete (Marek & Truszczyński, 1991) | thm-coNP-complete Th. 8 |

Table 2: Complexity of inference and Space Efficiency of Theorem Representations

Figures 3 and 4 contain the same information of Tables 1 and 2, but highlight existing reductions. Each figure contains two diagrams, the left one showing the existence of polynomial-time reductions among formalisms, the right one showing the existence of poly-size reductions. An arrow from a formalism to another denotes that the former can be reduced to the latter one. We use a bidirectional arrow to denote arrows in both directions and a dashed box to enclose formalisms that can be reduced one into another. Note that some formalisms are more appropriate in representing sets of models, while others perform better on sets of formulae. An interesting relation exists between skeptical default reasoning and circumscription. While there is no model-preserving poly-size reduction from circumscription to skeptical default reasoning (Gogic et al., 1995), a theorem-preserving poly-size reduction exists, as shown by Theorem 14.

## 7. Related Work and Conclusions

The idea of comparing the compactness of KR formalisms in representing information is not novel in AI. It is well known that first-order circumscription can be represented in second-order logic (Schlipf, 1987). Kolaitis and Papadimitriou (1990) discuss several computational aspects of circumscription. Among many interesting results they show a reduction from a restricted form of first-order circumscription into first-order logic. The proposed reduction will increase the size of the original formula by an exponential factor. It is left as an open problem to show whether this increase is intrinsic, because of the different compactness properties of the two formalisms, or there exists a more space-efficient reduction. When a

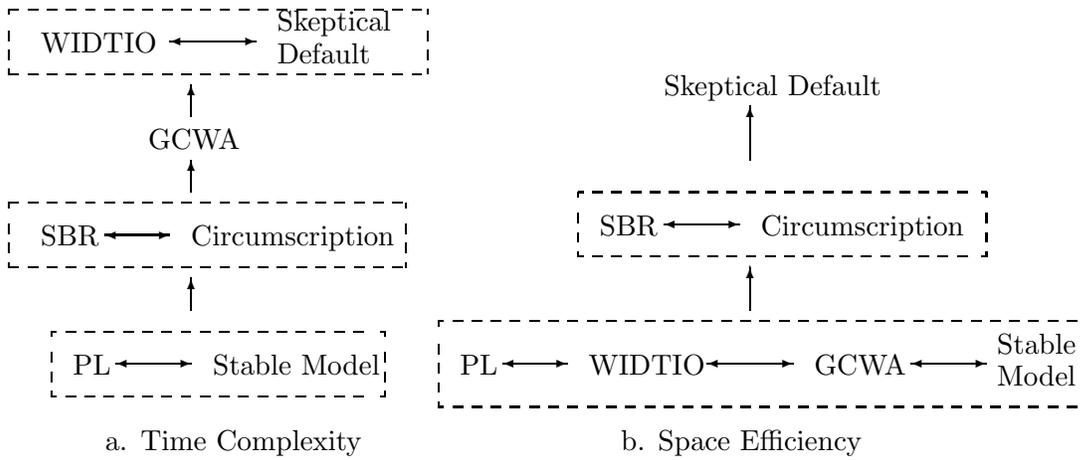

a. Time Complexity      b. Space Efficiency

Figure 3: Complexity of Model Checking vs. Space Efficiency of Model Representation

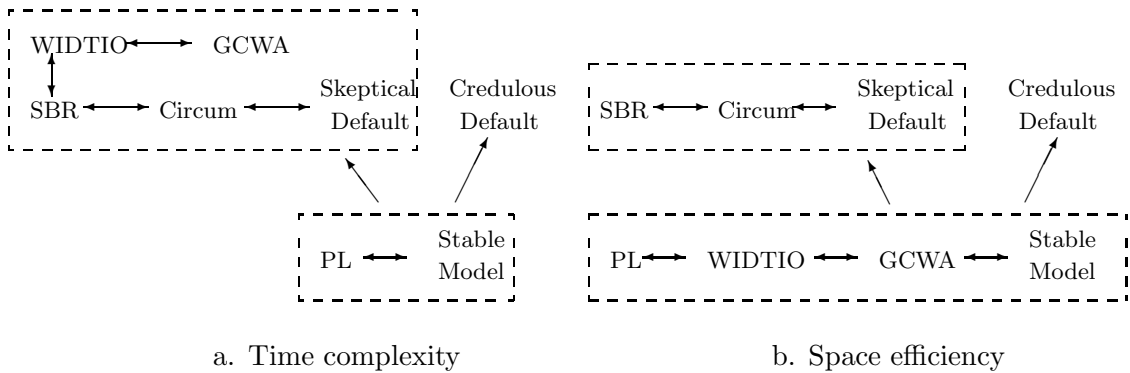

a. Time complexity      b. Space efficiency

Figure 4: Complexity of Inference vs. Space Efficiency of Theorem Representation

first-order language is used, more results on compactness and existence of reductions are reported by Schlipf (1995).

Khardon and Roth (1996, 1997), and Kautz, Kearns and Selman (1995) propose model-based representations of a $KB$ in Propositional Logic, and compare it with formula-based representations. Although their results are significant for comparing representations within PL, they refer only to this formalism, hence they are not applicable to our comparison between *different* PKR formalisms. The same comment applies also to the idea of representing a $KB$ with an efficient basis by Moses and Tennenholz (1996), since it refers only to one PKR formalism, namely, PL.

An active area of research studies the connections of the various non-monotonic logics. In particular, there are several papers discussing the existence of translations that are polynomial in time and satisfy other intuitive requirements such as *modularity* and *faithfulness*. Janhunen (1998), improving on results of Imielinski (1987) and Gottlob (1995), shows that default logic is the most expressive, among the non-monotonic logics examined, since both circumscription and autoepistemic logic can be modularly and faithfully embedded in default logic, but not the other way around. While these results are of interest and help to fully understand the relation among many knowledge representation formalisms, they are not directly related to ours. In fact, we allow for translations that are more general than polynomial time, while in all of the above papers they only consider translations that use polynomial time and also satisfy additional requirements.

The first result on compactness of representations for a propositional language is presented, to the best of our knowledge, by Kautz and Selman (1992). They show that, unless there is a collapse in the polynomial hierarchy, the size of the smallest representation of the least Horn upper bound of a propositional theory is superpolynomial in the size of the original theory. These results are also presented in a different form in the more comprehensive paper (Selman & Kautz, 1996). The technique used in the proof has been then used by us and other researchers to prove several other results on the relative complexity of propositional knowledge representation formalisms (Cadoli et al., 1996, 1997, 1999; Gogic et al., 1995).

In a recent paper (Cadoli et al., 1996b) we introduced a new complexity measure, i.e., *compilability*. In this paper we have shown how this measure is inherently related to the succinctness of PKR formalisms. We analyzed PKR formalisms with respect to two succinctness measures: succinctness in representing sets of models and succinctness in representing sets of theorems.

The main advantage of our framework is the machinery necessary for a formal way of talking about the relative ability of PKR formalisms to compactly represent information. In particular, we were able to formalize the intuition that a specific PKR formalism provides "one of the most compact ways to represent models/theorems" among the PKR formalisms of a specific class.

In our opinion, the proposed framework improves over the state of the art in two different aspects:

1. All the proofs presented in the previous papers only compare pairs of PKR formalisms, for example propositional circumscription and Propositional Logic (Cadoli et al., 1997). These results do not allow for a precise classification of the level of compactness of the considered formalisms. Rephrasing and adapting these results in our framework allows us to infer that circumscription is model-coNP-complete and thm-$\Pi_2^p$-complete. As a consequence, we also have that it is more space-efficient of the WIDTIO belief revision formalism in representing sets of models or sets of theorems.

2. Using the proposed framework it is now possible to find criteria for adapting existent polynomial reductions showing C-hardness into reductions that show model-C or thm-C-hardness, where C is a class in the polynomial hierarchy (Liberatore, 1998).

## Acknowledgments

This paper is an extended and revised version of a paper by the same authors appeared in the proceedings of the fifth international conference on the principles of knowledge representation and reasoning (KR'96) (Cadoli, Donini, Liberatore, & Schaerf, 1996a). Partial supported has been given by ASI (Italian Space Agency) and CNR (National Research Council of Italy).